\documentclass[journal,transmag]{IEEEtran}
\usepackage{hyperref,color,breqn,multirow}
\usepackage[top=0.7in, left=0.65in]{geometry}
\usepackage{color}
\usepackage{graphicx}
\begin{document}
\title{\fontsize{17pt}{17pt}\selectfont Design of Induction Machines using \\ Reinforcement Learning}
\author{\IEEEauthorblockN{Yasmin SarcheshmehPour\IEEEauthorrefmark{1,2},
Tommi Ryyppö\IEEEauthorrefmark{1},
Victor Mukherjee\IEEEauthorrefmark{1}, and Alex Jung\IEEEauthorrefmark{2}}
\vspace{5pt}
\IEEEauthorblockA{\IEEEauthorrefmark{1} Technology Center, ABB Motors and Generators, 00380 Helsinki, Finland}
\IEEEauthorblockA{\IEEEauthorrefmark{2} Department of Computer Science, Aalto University, 02150 Espoo, Finland}}

\IEEEtitleabstractindextext{%
\begin{abstract}
The design of induction machine is a challenging task due to different electromagnetic and thermal constraints. Quick estimation of machine's dimensions is important in the sales tool to provide quick quotations to customers based on specific requirements. The key part of this process is to select different design parameters like length, diameter, tooth tip height and winding turns to achieve certain torque, current and temperature of the machine. Electrical machine designers, with their experience know how to alter different machine design parameters to achieve a customer specific operation requirements. 
We propose a reinforcement learning algorithm to design a customised induction motor.
The neural network model is trained off-line by simulating different instances of electrical machine design game
with a reward or penalty function when a good or bad design choice is made. The results demonstrate that the suggested method automates electrical machine design without applying any human engineering knowledge.\\
\end{abstract}

}

\maketitle
\thispagestyle{empty}
\pagestyle{empty}

\section{Introduction}
\IEEEPARstart{E}{lectric} motors design and dimension estimation need to be fast, especially in sales tool in big corporation where thousands of motors are manufactured daily  based on different customer requirements. This is specifically very important in large scale industry where customer ask for quick dimension of machines based on their application, efficiency, load curve and operational conditions. In this case, an electric motor designer select a machine which is closely matching with the customer requirements from an existing database, often referred as base machine, and then manually change few machine parameters to match the customer requirements. Normally, this is done with a deterministic approach either manually or with an algorithm where machine designer increase or decrease certain machine design parameters to check if the customer requirements are met. However, this is still a time consuming process, and can be an impediment to give customer an immediate feedback of machine design, price etc during a sales negotiation. 

Using meta-heuristic optimisation algorithm to resolve the above mentioned issue can be challenging as in electric machine design optimisation problem some fundamental variables have more importance than other geometrical shape variables. In \cite{sato_2021}, a reinforcement learning algorithm and evolutionary optimisation method is proposed to tackle this issue for a linear induction motor. However, finding fundamental machine variables like pole pair number, slot number with stator and rotor slot shape dimensions seems a fairly complex and time consuming process. Also, in \cite{sato_2021}, the proposed method design machine from scratch, whereas in our current context, the optimisation space is comparatively simpler as the base induction machine already have most of the correct fundamental machine design variables like pole numbers, stator and rotor slot numbers, machine diameters etc.  Due to the manufacturing limit of our case, only few fixed diameters of stator and rotor laminations and rotor slots dimensions can be selected. Upon selecting a base machine, 3 machine design variables (length of the machine, rotor tooth tip height and number of coil turns) need to be selected automatically to meet the customer specification. If the goal is not met, then next base machine is selected, for example, with a bigger diameter and other geometrical parameters, and then again the 3 machine design variables are searched to met the customer requirements.

Reinforcement Learning (RL) methods have been applied to optimization problems arising in important application domains such as 
arithmetic circuits, chip placement, aerodynamic design and so on. In all these applications a deep RL algorithm is chosen to solve the corresponding optimization problem  \cite{hui_2021}. In our context, an agent plays different games of electrical machine design and at each step it checks if the design choices has brought it closer to the objective. 
The usefulness of an action 
is quantified by some reward signal. After playing several games, the agent learns what kind of design choices like altering length, coil turns and tooth tip height need to be made to achieve certain torque speed curve, temperature rise, efficiency, weight of the machine etc. We propose a novel designing method for the induction machines based on RL \cite{sutton_rl_2018} which use proximal policy optimisation (PPO) \cite{schulman_2017} to update the policy neural network (NN) based on cumulative reward for every game. The developed neural network model decreases the machine designing computational time to less than a minute. 
To the best knowledge of the authors, 
this is the first time when a design automation method of rotating electrical machine is investigated using RL, Proximal Policy Optimization (PPO) and NN. 
This model is developed for sales tool, however, engineers can also achieve an excellent initial machine design.

\section{Problem Formulation}
\label{problem_formulation}

Designing an electrical machine can be formulated as a parameter optimization problem that consists of an objective function $F(x)$, design variables $x$ as well as equality and inequality constraints. 
In our simplified electrical machine design problem,  the feasibility of the machine design is checked with certain performance values which are rotor tooth tip height, airgap flux density, current, torque, and stator temperature rise. These performance values are directly or indirectly related to customer specific requirements and manufacturing constraints.
In practice, there is a performance flag $\in \{-1, 0, 1\}$ for each of these performance values that checks one characteristic of the electrical machine. For every flag, if its relevant performance value meets the requirements, then the flag value is $0$, otherwise it is either $+1$ or $-1$.  A machine design can be considered as feasible if and only if all its flags are equal to zero.
we model the above mentioned electrical machine design problem as a RL problem. We solve this RL problem using PPO. 

\section{Method}
\label{sec:method}

RL problems can be formulated as Markov Decision Processes (MDPs), consisting of three key elements, states, actions, and reward \cite{hui_2021}. 
In this setting, each state is the concatenations of the performance flags’ values, and the previous action.
In this paper, the action space consists of $6$ different actions, decreasing/increasing length,
number of coil turns, and rotor tooth tip height of the machine.
The agent (policy network) starts from an initial state $s_0$, which is a base machine being
calculated based on the user specified requirements and the manufacturing constraints, and the final state $s_T$ corresponds to a feasible electrical machine that fits the given set of requirements and constraints. 
At each time step $t$, the agent begins in state $s_t$, takes an action $a_t$ that changes the design variables, arrives at a new state ($s_{t+1}$), and receives a reward ($r_t$) that is calculated based on the performance values. 
Through repeated episodes (sequences of states, actions, and rewards), the policy network learns to take actions that maximizes the cumulative reward. We use a model-free on-policy RL method called PPO \cite{schulman_2017} to update the parameters of the policy network.
PPO algorithms are policy gradient methods, which implies that they search the space of policies rather than assigning values to state-action pairs. 
As the optimization direction is guided by cumulative rewards, the reward function is the key element of RL methods. 

In this paper, for designing the reward function we only consider the flags' values. 
Each flag's value ($+1$ or $-1$) indicates that the action should decrease or increase the corresponding performance value. 
Therefore, for the flag $f_i$, the related reward, $w_{f_i}$, is $w_p > 0$ (positive reward) as long as the corresponding performance value is modified towards its correct direction, otherwise it is $w_n < 0$ (negative reward). As an example, if $torque$ flag is $+1$,  the action that increases the $torque$ performance value gives a positive reward ($w_{torque} = w_p$) and the action that decreases it gives a negative reward ($w_{torque} = w_n$). For combining multiple objectives into one reward function, we take the sum of the rewards for all the flags. We continue giving these rewards and penalties until all the flags' values reach to $0$.

Sometimes agent can not find the correct machine design after getting stuck in the different loops. In order to prevent these loops, some priorities are defined among these flags. In addition to this, a negative penalty is given to the actions that leads to a machine state which has been already found during the training.
The agent also loses the game if it reaches a fixed maximum number of total steps. When the agent wins the game by getting all zero flags, then it receives a huge reward.

\section{Results}
\begin{figure}[h]
\centering
\includegraphics[width=5.5cm]{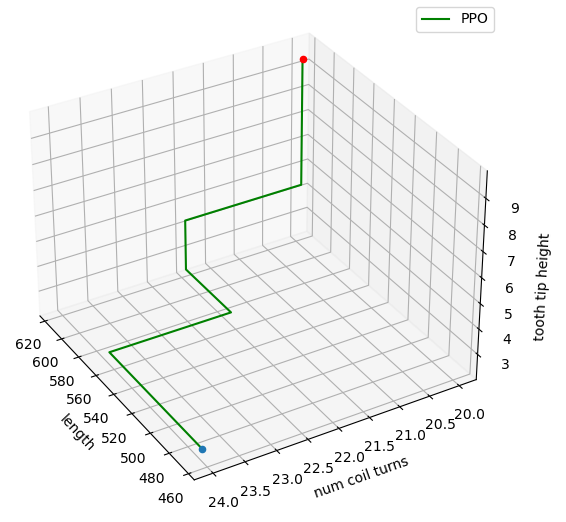}
\vspace{-0.2cm}
\caption{PPO steps for a single induction machine.}
\label{fig:fig_steps}
\end{figure}

Three different induction machines are selected as a case study for this work.
The PPO model is trained on 75 different variations of these selected induction machines.
The game is designed in a way that for each machine an initial design $s_0$ is calculated based on the power, voltage, and type of the machine. As mentioned in the section \ref{problem_formulation}, the observation state consists of the values of the design flags and the initial action. Then at each step $t$, one of the $6$ available actions 
is chosen. The maximum number of total steps is 300. Table \ref{flags_and_perfs} indicates the average number of steps for PPO for the selected machines. Figure \ref{fig:fig_steps} also represent PPO steps for a single induction machine. In the full paper, more details on the RL implementation, hyperparameter tuning, reward functions and its influence on the game's success and failure will be discussed. Also, more detail analysis on the design outcomes of different machines (different power, size etc) and scaling of the methods for different base machines will be presented.

\begin{table}[h]
\centering
\caption{The average number of steps for PPO.}
\label{flags_and_perfs}
 \begin{tabular}{ | p{1.0cm} |  p{1.1cm} | p{1.0cm} |p{1.2cm} |}
    \hline
    Machine & power & voltage & PPO steps \\ \hline
    \quad 1 & 2500 kW & 10000 V & 11 \\ \hline
    \quad 2 & 600 kW & 6000 V & 12 \\ \hline
    \quad 3 & 2100 kW & 6000 V & 5 \\ \hline
    \end{tabular}
\end{table}


\vspace{-0.5cm}


\begin{thebibliography}{1}

\bibitem{sato_2021}
T. Sato and M. Fujita, \emph{"A Data-Driven Automatic Design Method for Electric Machines Based on Reinforcement Learning and Evolutionary Optimization,"} in IEEE Access, vol. 9, pp. 71284-71294, 2021, doi: 10.1109/ACCESS.2021.3078668.



\bibitem{hui_2021}
Xinyu Hui, Hui Wang, Wenqiang Li, , Junqiang Bai, Fei Qin, and Guoqiang He , \emph{"Multi-object aerodynamic design optimization using deep reinforcement learning"}, AIP Advances 11, 085311 (2021) https://doi.org/10.1063/5.0058088


\bibitem{sutton_rl_2018}
R. Sutton, and A. Barto.  \emph{"Reinforcement Learning: An Introduction."} Second : The MIT Press, 2018 .


\bibitem{schulman_2017}
J. Schulman, F. Wolski, P. Dhariwal, A. Radford, and
O. Klimov. \emph{"Proximal policy optimization algorithms."} arXiv preprint arXiv:1707.06347 (2017).


\end{thebibliography}
\end{document}